\documentclass[acmtog,authorversion,nonacm]{acmart}

\usepackage{booktabs} 
\usepackage{multirow}
\usepackage{caption}
\usepackage{bm}

\newcommand\blfootnote[1]{%
  \begingroup
  \renewcommand\thefootnote{}\footnote{#1}%
  \addtocounter{footnote}{-1}%
  \endgroup
}

\usepackage[ruled]{algorithm2e} 

\SetAlFnt{\small}
\SetAlCapFnt{\small}
\SetAlCapNameFnt{\small}
\SetAlCapHSkip{0pt}

\begin{document}

\begin{teaserfigure}
\includegraphics[width=7.0in]{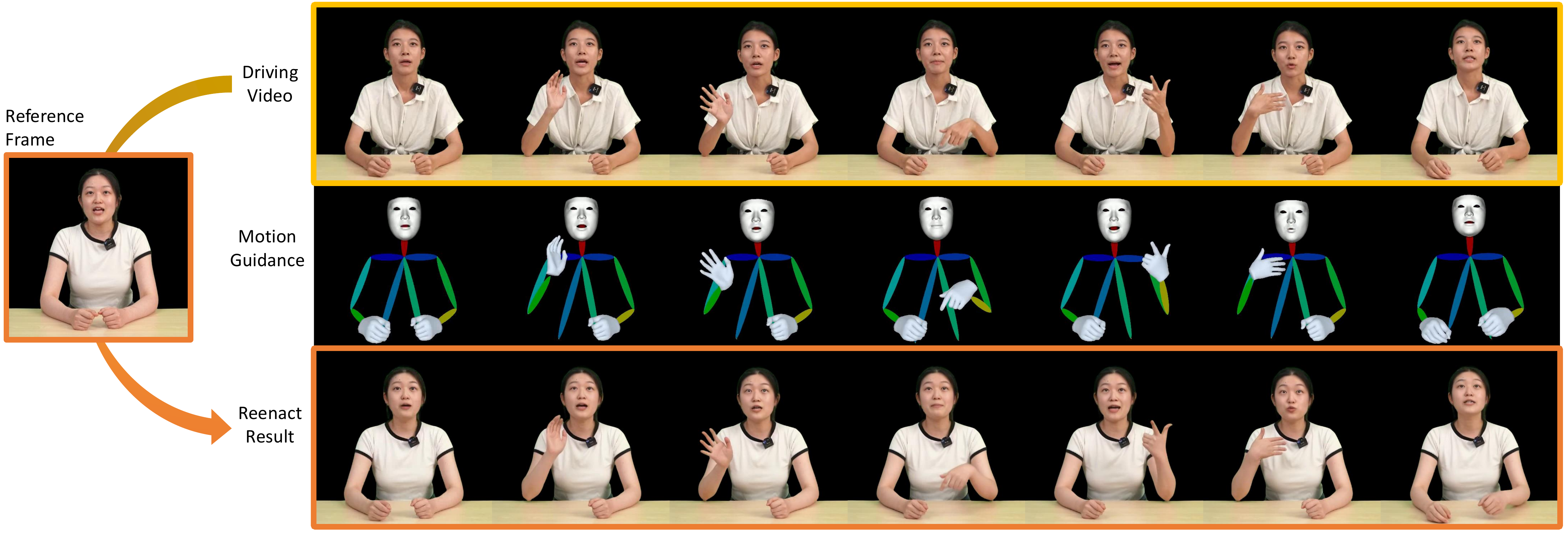}
\caption{
\textbf{TALK-Act.} 
Our method enables temporal-consistent high-fidelity avatar reenactment from short footage of a single-view video.
}
 \label{fig:teaser}
\end{teaserfigure}

\title{TALK-Act: Enhance Textural-Awareness for 2D Speaking Avatar Reenactment with Diffusion Model}

\author{Jiazhi Guan}
\authornotemark[1]
\email{guanjz20@mails.tsinghua.edu.cn}
\affiliation{%
  \institution{DCST, BNRist, Tsinghua University}
  \country{China}
}
\author{Quanwei Yang}
\email{yangquanwei@mail.ustc.edu.cn}
\affiliation{%
  \institution{University of Science and Technology of China}
  \country{China}
}
\author{Kaisiyuan Wang}
\email{wangkaisiyuan@baidu.com}
\affiliation{%
  \institution{Department of Computer Vision Technology (VIS), Baidu Inc.}
  \country{China}
}
\author{Hang Zhou}
\authornotemark[2]
\email{zhouhang09@baidu.com}
\affiliation{%
  \institution{Department of Computer Vision Technology (VIS), Baidu Inc.}
  \country{China}
}
\author{Shengyi He}
\email{heshengyi@baidu.com}
\affiliation{%
  \institution{Department of Computer Vision Technology (VIS), Baidu Inc.}
  \country{China}
}
\author{Zhiliang Xu}
\email{xuzhiliang@baidu.com}
\affiliation{%
  \institution{Department of Computer Vision Technology (VIS), Baidu Inc.}
  \country{China}
}
\author{Haocheng Feng}
\email{fenghaocheng@baidu.com}
\affiliation{%
  \institution{Department of Computer Vision Technology (VIS), Baidu Inc.}
  \country{China}
}
\author{Errui Ding}
\email{dingerrui@baidu.com}
\affiliation{%
  \institution{Department of Computer Vision Technology (VIS), Baidu Inc.}
  \country{China}
}
\author{Jingdong Wang}
\email{wangjingdong@outlook.com}
\affiliation{%
  \institution{Department of Computer Vision Technology (VIS), Baidu Inc.}
  \country{China}
}
\author{Hongtao Xie}
\email{htxie@ustc.edu.cn}
\affiliation{%
  \institution{University of Science and Technology of China}
  \country{China}
}
\author{Youjian Zhao}
\authornotemark[2]
\email{zhaoyoujian@tsinghua.edu.cn}
\affiliation{%
  \institution{DCST, BNRist, Tsinghua University and Zhongguancun Laboratory}
  \country{China}
}
\author{Ziwei Liu}
\email{}
\affiliation{%
  \institution{S-Lab, Nanyang Technological University}
  \country{Singapore}
}

\authorsaddresses{}

\begin{abstract}

Recently, 2D speaking avatars have increasingly participated in everyday scenarios due to the fast development of facial animation techniques. However, most existing works neglect the explicit control of human bodies. In this paper, we propose to drive not only the faces but also the torso and gesture movements of a speaking figure. Inspired by recent advances in diffusion models, we propose the Motion-Enhanced \textbf{T}extural-\textbf{A}ware Mode\textbf{L}ing for Spea\textbf{K}ing \textbf{A}vatar Reena\textbf{ct}ment (\textbf{TALK-Act}) framework, which enables high-fidelity avatar reenactment from only short footage of monocular video. Our key idea is to \emph{enhance the textural awareness with explicit motion guidance} in diffusion modeling. Specifically, we carefully construct 2D and 3D structural information as intermediate guidance. While recent diffusion models adopt a side network for control information injection, they fail to synthesize temporally stable results even with person-specific fine-tuning. We propose a Motion-Enhanced Textural Alignment module to enhance the bond between driving and target signals. Moreover, we build a Memory-based Hand-Recovering module to help with the difficulties in hand-shape preserving. After pre-training, our model can achieve high-fidelity 2D avatar reenactment with only 30 seconds of person-specific data. Extensive experiments demonstrate the effectiveness and superiority of our proposed framework.
Resources can be found at \url{https://guanjz20.github.io/projects/TALK-Act}.
\end{abstract}


\maketitle

\blfootnote{
$*$ Work done during an internship at Baidu Inc. \quad$\dag$ Corresponding author.
}

\section{Introduction}
\label{sec:intro}

The creation of 2D avatars has become increasingly pivotal to various fields. In particular, lifelike speaking avatars derived from monocular cameras have seamlessly integrated into everyday life, serving as virtual assistants, TV anchors, reporters, salespeople, and even teachers.
Alongside the needs, great efforts have been devoted to 2D avatar animation over the years~\cite{kim2018deep,zhou2019talking,kaisiyuan2020mead,burkov2020neural,ji2022eamm,ji2021audio,lu2021live,liu2022semantic,zheng2022avatar,zielonka2023insta,cheng2022videoretalking,thies2020neural,prajwal2020lip,siarohin2019first,wang2022latent} with a specific emphasis on human faces.

However, presenting only head dynamics does not satisfy the needs of real-world scenarios where avatars with natural gesture movements are perceivably more realistic. However, most gestures of 2D avatars are limited to the recordings.
Thus, it is highly desirable that the whole figure can be explicitly controlled at high fidelity. Previous methods propose to use warping-based methods~\cite{siarohin2021motion,zhao2022thin,wang2022latent,siarohin2023unsupervised}, GANs~\cite{chan2019everybody,liu2019liquid,kappel2021high}, and neural rendering~\cite{jiang2022neuman} to drive the whole body. However, these methods lead to obvious artifacts on limbs and faces due to the limited modeling ability, making them not applicable under most circumstances.

More recently, studies with diffusion models~\cite{ho2020denoising,rombach2022high,song2020denoising} have explored whole-body control and reenactment under the one-shot settings~\cite{zhang2023adding,mou2024t2i,guo2023animatediff,chang2023magicpose,hu2023animate,xu2023magicanimate,zhu2024champ} with different human-related structural conditions like 2D and 3D poses~\cite{yang2023effective,guler2018densepose}, 3D paremtric human models~\cite{SMPL:2015,SMPL-X:2019}, and depth~\cite{guler2018densepose} as guidance. With the well-trained Stable Diffusion~\cite{rombach2022high}, explicit controls can be achieved by involving a \emph{reference or control network} for integrating additional control information. 
Their results are notably unstable due to insufficient use of appearance information and a mismatch between their training protocol and this task.
Concurrently, Make-Your-Anchor~\cite{huang2024make} adopts a two-stage training protocol with generalized pretraining and personalized fine-tuning. They leverage rendered 3D models and a reference frame as conditions and send them to a ControlNet~\cite{zhang2023adding} branch as driving signals. To bolster facial identity recovery, an additional face-enhancement diffusion model is integrated into the network. Though appealing anchor-style avatars can be created, they cannot faithfully build the relationship between detailed textures and their control signal. Thus this method cannot fully express the target pose and render visible textural flaws.

In this paper, we propose the Motion-Enhanced \textbf{T}extural-\textbf{A}ware Mode\textbf{L}ing for Spea\textbf{K}ing \textbf{A}vatar Reena\textbf{ct}ment (\textbf{TALK-Act}) framework, which achieves high-fidelity 2D
speaking avatars reenactment between monocular videos. Our key insight is to \emph{enhance the textural awareness with explicit motion guidance} in diffusion modeling. We construct the structural motion guidance from three different sources: 2D pose, 3D rendering of parametric faces, and 3D
 rendering of hands, to ensure model accuracy and processing efficiency. A two-branch design with a denoising main network and a Reference Branch is adopted. 
Similar to~\cite{huang2024make}, we rely on relatively large-scale academic data for learning generalized rendering capability, 
then build mappings between structural motion guidance with personalized textural by person-specific tuning. 
However, previous methods~\cite{huang2024make,hu2023animate} integrate textural and structural information within vanilla cross-attention, making them unable to successfully express person-specific textures. They either lose textural temporal consistency or overfit the motion of a specific person, both leading to artifacts given unseen driving poses.

In our work, we devise a simple  \emph{Motion-Enhanced Textural Alignment Module}. It aligns textural appearance information according to the spatial correspondence between the reference and driving structural guidance. 
The correspondence explicitly enforces textural awareness of the denoising network from two different perspectives: the main branch's input, and its interaction with the Reference Branch. With enhanced textural awareness, our model generates highly consistent results even when it is refined on videos less than 30 seconds.
Furthermore, synthesizing detailed hand motions has been challenging. We propose a \emph{Memory-based Hand-Recovering Module} by enhancing hand textures with a learnable hand memory bank. Experiments show that our method reenacts 2D speaking avatars with better fidelity and stability than the previous state-of-the-art methods. 

Our contributions can be summarized as follows: \textbf{1)} We propose the \textbf{TALK-Act} framework which improves the overall avatar reenactment quality with our proposed \emph{Motion-Enhanced Textural Alignment Module}. \textbf{2)} We propose a \emph{Memory-based Hand-Recovering Module} that tackles the difficulty of synthesizing stable hand movements. \textbf{3)} Extensive experiments demonstrate that our model clearly outperforms previous methods. Moreover, our method simultaneously supports audio-driven facial reenactment and few-shot tuning on only 30-second videos of target identities.

\section{Related Work}
\label{sec:related}

\subsection{Talking Head Generation}
A significant number of studies on 2D avatars have been conducted to generate realistic talking heads~\cite{zhou2021pose,chen2019hierarchical,chen2020comprises,guo2021adnerf, ji2022eamm, wang2021audio2head, song2018talking, ma2023styletalk,ma2023talkclip,wang2022pdfgc,wang2021audio2head,yu2022thpad,zhang2023sadtalker,guan2024resyncer}, which is not realistic enough without a body. 
Certain studies \cite{chen2018lip,prajwal2020lip, thies2020neural, song2018talking, park2022synctalkface, sun2022masked, cheng2022videoretalking, guan2023stylesync,wu2023speech2lip} directly edit mouth shapes based on input audio, achieving accurate lip-sync and high-fidelity results.
Despite advancements in various methods for generating talking heads, these approaches are limited by their focus areas and are unable to convey human body movements that are essential for vivid expression.

\subsection{Pose-Guided Human Body Animation}
Earlier attempts to generate full-body movements are mostly achieved by motion estimation. These methods assume a coherent pose motion with small spatial deformation in target videos. Flows are estimated from detected keypoints from adjacent frames to express movements in~\cite{DBLP:journals/corr/abs-1804-07739,liu2019liquid,ren2020deep}. Moreover, FOMM~\cite{siarohin2019first} proposes to implement motion transfer conditioned on estimated sparse keypoints and local affine transformation. Though more methods are proposed based on a similar pipeline~\cite{zhao2022thin,siarohin2021motion,wang2022latent},
limited by the capacity of earlier generative models, these methods cannot support high-fidelity generation and inevitably produce noticeable artifacts. This stands the same for GAN-based methods with 2D and 3D representations~\cite{chan2019everybody,liu2019liquid}

Recent studies leverage the strong Diffusion-based models for tackling this task.
A line of studies is conducted based on a UNet-based network~\cite{ronneberger2015u} with information injection by cross-attention~\cite{vaswani2017attention}. 
PIDM~\cite{bhunia2023person} first introduces classifier-free diffusion guidance to achieve pose-guided human image generation~\cite{liao2024appearance}.
DreamPose~\cite{karras2023dreampose} uses UV maps as motion signals and performs conditional embeddings to achieve motion transfer.
Similar ideas are also explored in \cite{wang2023disco,zhu2024champ,xu2023magicanimate,hu2023animate}. 
These methods employ the powerful Stable-Diffusion~\cite{rombach2022high} as the generative backbone and insert temporal layers to increase the overall motion smoothness of generated videos.
Typically, the denoising process of the backbone is conditioned on motion features,
which could be encoded by a ControlNet~\cite{zhang2023adding} or a conv-based feature encoder from various types of pose input, e.g., body skeleton detected by~\cite{cao2017realtime}.
Though plausible results have been delivered, these methods show capacity only on coarse-grained human movement generation and struggle to maintain stable and consistent high-fidelity outcomes on detailed regions like faces and hands. 
A recent study~\cite{huang2024make} proposes a two-stage diffusion pipeline to achieve fine-grained anchor generation. 
Their method cannot correctly restore the textural information and occasionally produces unsatisfactory facial expressions and hand gestures.

\section{Methodology}
\label{sec:method}
\begin{figure*}[!t]
\centering
\includegraphics[width=\linewidth]{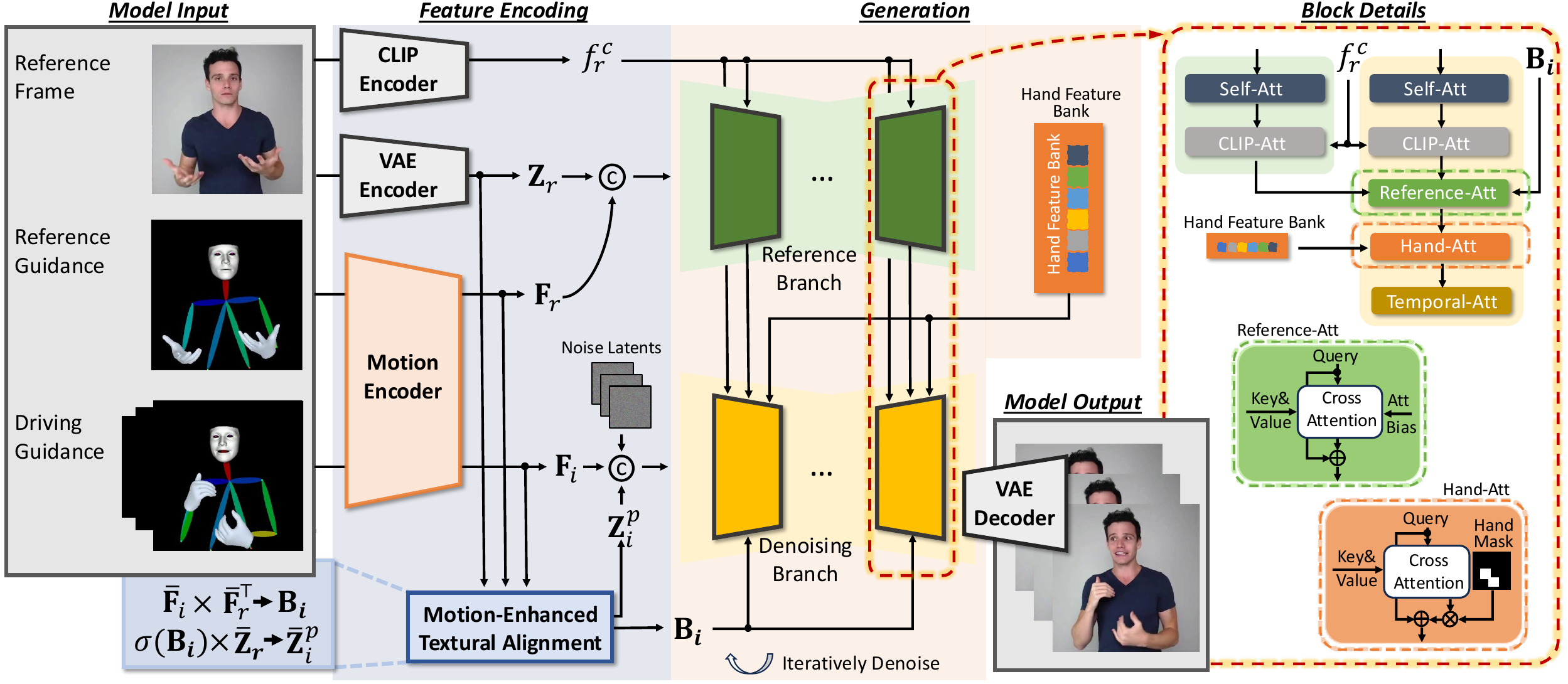}
\caption{
\textbf{TALK-Act Framework}. 
The framework is depicted in five parts:
1) \textit{{Model Input}}. The model input is shown on the left side.
2) \textit{Feature Encoding}. Initial feature encoding is handled by three encoders and the proposed Motion-Enhanced Textural Alignment module.
3) \textit{Generation}. Two branches, the Reference Branch and Denoising Branch, take input features to generate pose-aligned frames through the denoising process.
4) \textit{Model Output}. The RGB outputs are then given by a decoder.
5) \textit{Block Details}. We elaborate on detailed structures of the designed UNet layers.
The driven subject is from \copyright \emph{Charisma on Command}.
}
\label{fig:pipeline}
\end{figure*}

An overall description of our proposed TALK-Act framework is illustrated in Fig.~\ref{fig:pipeline}. 
We first introduce the formulation of our task and review the preliminaries of diffusion models in Sec.~\ref{sec:preliminary}. 
Then we introduce the pipeline of our TALK-Act framework and its newly proposed components in Sec.~\ref{sec:method_framework}.

\subsection{Task Formulation and Preliminaries}
\label{sec:preliminary}

\paragraph{Task Description.} 
This work tackles the problem of 2D speaking avatar reenactment, which transfers the complete motion information including the movements of body poses, facial expressions, and hand gestures, from one driving person to another target identity as shown in Fig.~\ref{fig:teaser}.

\paragraph{Structural Guidance.} Given the intricate nature of motion information, adopting structural guidance as an intermediary step can mitigate the learning challenges.
Recent studies~\cite{hu2023animate,xu2023magicanimate,huang2024make} leverage various motion signals including 3D meshes from SMPL-X~\cite{SMPL-X:2019}, 2D skeletons from DWPose~\cite{yang2023effective}, 2D gesture maps from DensePose~\cite{guler2018densepose}, etc. However, 2D skeletons or gesture maps can only provide sparse and coarse structural guidance for delicate local regions like the face and hands, leading to missing details.
On the other hand, though 3D fitting results provide compact geometric guidance, they suffer from temporal inconsistency.

\begin{figure}[t]
\centering
\includegraphics[width=0.9\linewidth]{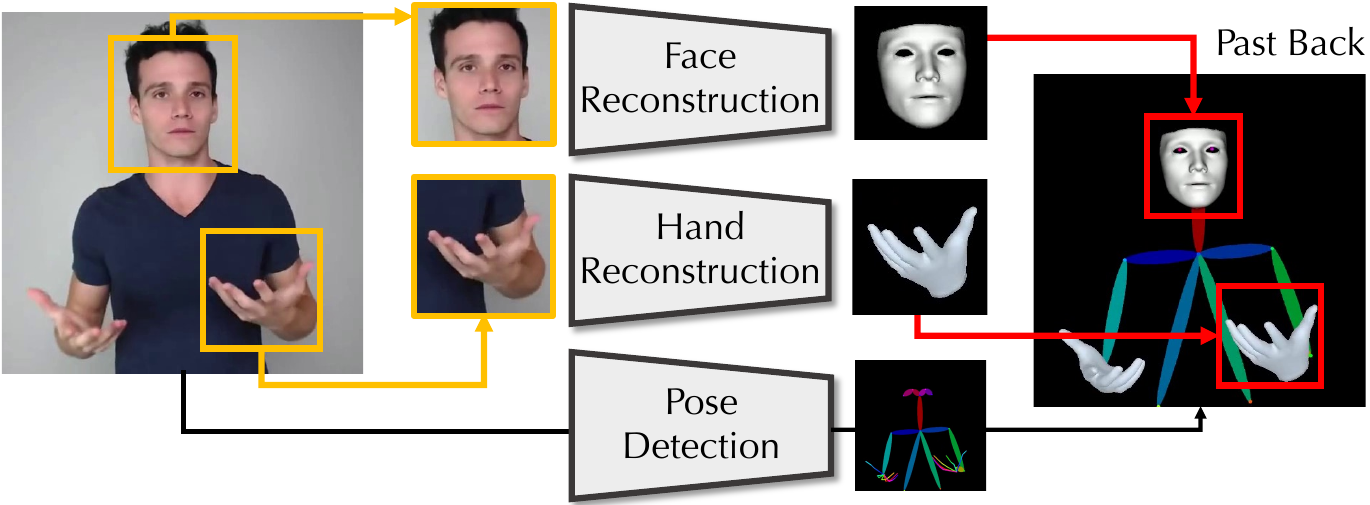}
\caption{
\textbf{Motion Guidance}. 
We illustrate how to create a specific structural guidance from an RGB frame. 
The subject is from \copyright \emph{Charisma on Command}.
}
\label{fig:motion_guidance}
\end{figure}

Instead, we construct the guiding signal by combining 2D and 3D representations. 2D skeletons from DWPose~\cite{yang2023effective}, 3D facial mesh from Deep3DReconstruct~\cite{deng2019accurate}, and 3D hand mesh from HaMeR~\cite{pavlakos2024reconstructing} are composed and rendered together as shown in Fig.~\ref{fig:motion_guidance}. The \emph{motion guidance} $\textbf{M}$ is thus presented in this form.

\paragraph{Training and Inference Formulation.} 
The training is performed under a self-reconstruction protocol.
Given a $T$-frame training video clip $\textbf{V} = \{{I}_{{1}}, \dots, {I}_{{T}}\}$ whose structural \emph{motion guidance} of each frame can be represented by $\textbf{M} = \{m_{1}, \dots, m_{T}\}$, the training goal is to recover the original frames $\textbf{V}$ with the driving guidance $\textbf{M}$ and a reference frame $I_r \in \textbf{V}$.

During inference, another driving video of a different identity $\textbf{V}' = \{I'_{{1}}, \dots, I'_{{T'}}\}$, 
with motion guidance $\textbf{M}' = \{m'_{1}, \dots, m'_{T'}\}$ is provided.
The goal is to transfer the driving motion to the target person and synthesize $\hat{\textbf{V}} = \{{\hat{I}}_{{1}}, \dots, {\hat{I}}_{{T'}}\}$ conditioning on a reference frame $I_r \in \textbf{V}$ and $\textbf{M}'$. The person in $\hat{\textbf{V}}$ should move in the same way as $\textbf{V}'$ with the appearance of $\textbf{V}$. Note that motion signals are aligned with each identity based on the shoulder lengths and positions. 
The 3D facial identity coefficients are also aligned with the target individual by substituting the identity coefficients with those of the driven identity, thereby ensuring identity-consistent guidance.

\paragraph{Diffusion Model Preliminaries.} 
Our framework is built upon the famous Stable Diffusion~\cite{rombach2022high}, which employs a Variational Autoencoder (VAE) with encoder $\bm{E}_{VAE}$ and decoder $\bm{D}_{VAE}$ for data compression and a UNet as the Denoising Network. Given an input image $I$, it is first encoded to $\textbf{Z}_{(0)} = \bm{E}_{VAE}(I)$. 
During the diffusion process, Gaussian noise $\epsilon_t$ perturbs $\textbf{Z}_{(0)}$ into a noisy latent $\textbf{Z}_{(t)}$ at each timestep $t$. The $(t)$ denotes that the feature map has been processed with $t$-step noise perturbation. They are neglected when simply denoting features encoded by the VAE encoder. The denoising UNet learns to reverse the procedure. The objective is defined as follows:
\begin{equation}
\mathcal{L}_t = \mathbb{E}_{\textbf{Z}_{(t)}, \textbf{c}, \epsilon_t, t}\left(\left\|\epsilon_t-\epsilon_\theta\left(\textbf{Z}_{(t)}, \textbf{c}, t\right)\right\|_2^2\right),
\end{equation}
where $\textbf{c}$ denotes the conditions.
In our case, the conditions are the reference image and the structural motion guidance.

During the inference phase, a random Gaussian distribution initializes $\textbf{Z}_{(t)}$, which is progressively denoised by $\epsilon_\theta$ to recover $\hat{\textbf{Z}}_{(0)}$. The denoising UNet iteratively estimates the noise component corresponding to each timestep $t$. Ultimately, $\hat{\textbf{Z}}_{(0)}$ is reconstructed into a real image $\hat{I}$ by the Decoder $\bm{D}_{VAE}$.

\subsection{Enhanced Textural-Aware Framework Designs}
\label{sec:method_framework}
\subsubsection{Two-Branch Framework Backbone.} 
\label{sec:3.2.1}
As mentioned in Sec. \ref{sec:intro}, incorporating a side network alongside the primary denoising backbone has been proven effective for injecting control information~\cite{zhang2023adding,huang2024make,hu2023animate,zhu2024champ}.

In our work, we adopt a similar two-branch protocol. Basically, we devise a Reference Branch that shares the block design and layer number with the main Denoising Branch.
Different from the design of ControlNet~\cite{zhang2023adding} and the concurrent study~\cite{huang2024make},
we deliver the driving motion guidance $\textbf{M}$ to the Denoising Branch so that the spatial information can be enhanced through skip connections. 
Then, it is natural to rely on the Reference Branch to process textural information of the reference frame $I_r$. The two branches interact with each other through cross-attention mechanisms. A CLIP~\cite{radford2021learning} feature $f^c_r$ encoded from the reference frame also interacts with the two branches with the cross-attention operation, which benefits the network training and appearance recovery.

However, such a vanilla design described above does not guarantee accurate Query-Key alignment in the cross-attentions to faithfully express textural information in the reference frame. Thus we identify three important aspects that would benefit the learning process: 
\textbf{1)} The injected information should be coherent between the Reference Branch and the Denoising Branch so that attention can be carried out with better effectiveness. 
\textbf{2)} The corresponding textural information aligned with the structural guidance should be highlighted.
\textbf{3)} Local regions with delicate details need specific treatment.

\subsubsection{Motion-Enhanced Textural Alignment Module}
\label{sec:3.2.2}

Given the insights from the previous analysis,  we propose to 
unify the injected representations with the aid of the reference frame's structural motion information, a natural bridge that has been neglected in previous studies.
While it is difficult to directly align the driving motion $\textbf{M}$ with textures, it would be much easier to build a correlation between two motion's structural guidance. Thus $m_r$ of the reference frame $I_r$ is also constructed.

\paragraph{Building Motion Correspondance.}
Specifically, a lightweight motion encoder $\bm{E}_{M}$, constructed using several conv-downsample layers, is employed.
The reference motion's feature map
$\textbf{F}_r \in \mathbb{R}^{h \times w \times c_0}$ is encoded from $\bm{E}_{M}(m_r)$, and the features of the whole driving motion sequence $\textbf{M} = \{ m_1, \dots, m_T \}$ is encoded to $\textbf{F} = \{\textbf{F}_1, \dots, \textbf{F}_T\}$ by $\bm{E}_{M}$ as well, where 
$\textbf{F}_i \in \mathbb{R}^{h \times w \times c_0}$ is the $i$th driving guidance's feature and $c_0$ is the channel dimension. The correspondence matrix between the driving and reference can be built by flattening these two kinds of feature maps to $\overline{\textbf{F}}_r \in \mathbb{R}^{(hw) \times c_0}$ and $\overline{\textbf{F}}_i \in \mathbb{R}^{(hw) \times c_0}$ and performing the query-key scaled dot product:
\begin{equation}
\label{eq:B}
    \textbf{B}_i = \frac{\overline{\textbf{F}}_i  {\overline{\textbf{F}}^\top_r}}{\sqrt{c_0}}.
\end{equation}
The correspondence matrix $\textbf{B}_i$ represents the similarity correlation between the reference frame and the $i$th driving motion at each spatial location. This matrix is subsequently utilized to enhance the textural awareness of the Denoising Branch, leveraging information for both the network input and the cross-attention modules.

\begin{figure}[t]
\centering
\includegraphics[width=0.9\linewidth]{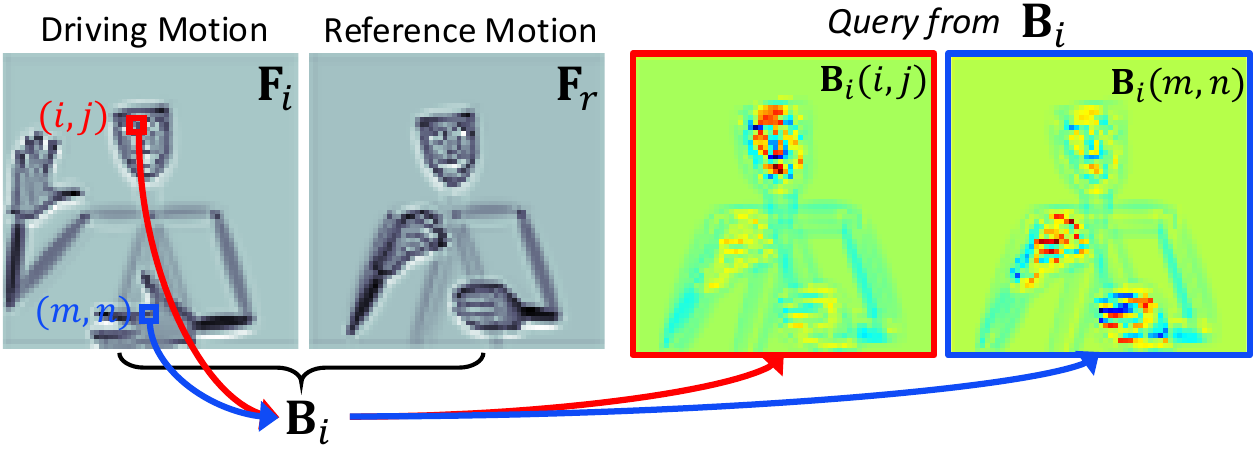}
\caption{
\textbf{Visualization of Corresponding Matrix}. 
From two positions of the driving motion, the two queried similarity heatmaps highlight a strong focus on the face and hand regions of the reference, respectively.
}
\label{fig:wapring}
\end{figure}

As depicted in Fig.~\ref{fig:wapring}, the correspondence matrix $\textbf{B}_i$ serves as an explicit indicator and attends to the corresponding regions.

\paragraph{Texturally Enhanced and Aligned Network Inputs.}
We first unify the format of the two branches' inputs. We argue that both structural and textural information should serve as the input of both branches for consistent information learning. Thus the input of the Reference Branch is the concatenation of the appearance feature $\textbf{Z}_r$ encoded from $\bm{E}_{VAE}(I_r)$ and the motion feature $\textbf{F}_r$.

On the other hand, it would be more beneficial when a texturally aligned 
pseudo feature map ${\textbf{Z}^{p}_i}$ can serve as the input of the Denoising Branch along the driving motion sequence. It would enrich the input format and provide explicit textural guidance.  
With the help of the correspondence matrix $\textbf{B}_i$ above, we thus derive the aligned appearance features for each $m_i$ by completing the cross-attention in Eq.~(\ref{eq:03}):
\begin{equation}
\label{eq:03}
    \overline{\textbf{Z}}^p_i = \text{softmax} (\textbf{B}_i) \overline{\textbf{Z}}_r,
\end{equation}
where $\overline{\textbf{Z}}^p_i$ and $\overline{\textbf{Z}}_r$ are features of the flattened form. Each aligned appearance feature ${\textbf{Z}}^p_i$ is later concatenated with the motion feature map $\textbf{F}_i$.

During training, the Denoising Branch processes a total of $n < T$ motion guidance frames in one pass with the current step noise $\textbf{Z}_{i(t)}$. Thus the final input can be represented as a concatenation of $\{\textbf{F}_{i:i+n}, {{\textbf{Z}}}^p_{i:i+n}, \textbf{Z}_{i:i+n(t)}\}$.

\paragraph{Texturally Enhanced Reference Cross-Attention.}
Furthermore, the correspondence matrix could benefit the cross-attention module between the Reference Branch and the Denoising Branch. We enhance the correlation between any query feature $Q$ from the Denoising Branch and the key, value feature $K$, $V$ from the reference with $\textbf{B}_i$ in the form of biased attention. The computation of the cross-attention \emph{Reference-Att}
 can be written as:
\begin{equation}
    \text{Reference-Att}(Q, K, V, \textbf{B}_i) = \text{softmax}\left(\frac{Q K^\top}{\sqrt{c}} + \textbf{B}_i\right) V,
\end{equation}
where $c$ is the corresponding channel dimension. 
Thus, the correspondence matrix $\textbf{B}_i$, which captures the similarity correlation between the reference and driving signals, will provide alignment priors for the cross-attention between the two branches.

\paragraph{Memory-based Hand-Recovering.} 
It is well-known that human hands are difficult to synthesize~\cite{rombach2022high}. We argue that relying only on the aligned hand features is not sufficient for recovering hand details. Thus we propose a global memory bank for restoring hand information under different scenes.

This is achieved by constructing a set of learnable parameters, referred to as the Hand Feature Bank $\textbf{H} \in \mathbb{R}^{N_b \times c}$, and incorporating \textit{Hand-Att} layers within the UNet architecture, where $N_b$ is the bank length and $c$ is the channel number.
The Hand Feature Bank $\textbf{H}$ can implicitly restore human hand textures during training with reconstruction supervision, and it provides reliable textural information according to similarity calculation. This bank $\textbf{H}$ serves as Key and Value in the following definition:
\begin{equation}
    \text{Hand-Att}(Q, \textbf{H}, Mask) = Mask \cdot \text{softmax}\left(\frac{Q \textbf{H}^\top}{\sqrt{c}}\right) \textbf{H},
\end{equation}
where $Q$ represents the query feature from the Denoising Branch. $Mask$ is the hand region mask built from the hand mesh, facilitating \textit{Hand-Att} to focus on the hand areas. The results of the hand attention are integrated into the feature maps with skip connections.

\paragraph{Temporal Modeling.} As the Denoising Branch processes a total of $n$ driving motions at once, we follow previous studies~\cite{guo2023animatediff,hu2023animate} to involve the temporal attention module 
\emph{Temporal-Att} 
into our work. Given any feature map $\textbf{A} \in \mathbb{R}^{b \times n \times h \times w \times c} $  with batch size $b$ of the Denoising network, we reshape it to $\textbf{A} \in \mathbb{R}^{(bhw) \times n \times c} $ and performs self-attention across the $n$ features. 
Their two-stage training protocol has also been adopted, please find a related discussion in the supplementary files.

Experiments validate that with the common temporal design, our consecutive alignment of textural features naturally enables textural stableness when synthesizing consecutive frames.

\section{Experiments}
\label{sec:experiment}

\begin{figure*}[!t]
\centering
\includegraphics[width=\linewidth]{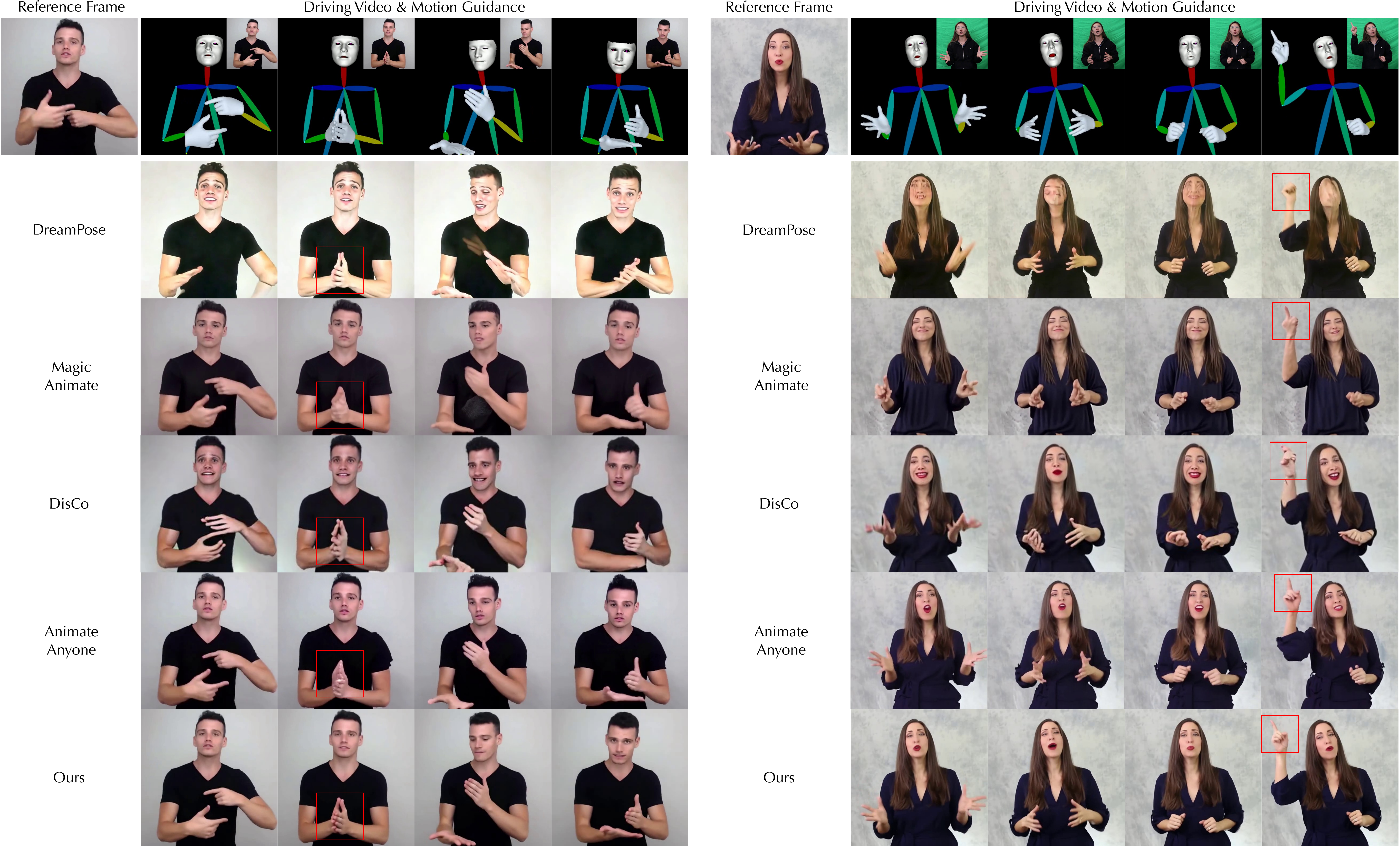}
\caption{
\textbf{Qualitative Comparisons}. 
We compare SOTA methods on both self-driven (left) and cross-driven (right) settings. 
All methods utilize ``Reference Frame'' for appearance reconstruction, depicted in the first row. 
Driving video and motion guidance are shown in the first row as well. 
The driving signals of other methods are omitted here. Please zoom in for a better visualization of animation details.
The driven subject on the left is from \copyright \emph{Charisma on Command}, and the driven subject on the right is from \copyright \emph{Vanessa Van Edwards, Science of People}.
}
\label{fig:cmp_sota}
\end{figure*}

\begin{table*}[]
\centering
\caption{\textbf{Quantitative Comparisons}. We show the best in bold. Methods with $^*$ are evaluated on the test set of the subject ``Seth''~\cite{ahuja2020style} only.}
{\scriptsize
\resizebox{\linewidth}{!}{
\begin{tabular}{l|ccccc|ccc|ccc}
\toprule
\multirow{2}{*}{Method} & \multicolumn{5}{c|}{\textit{Visual Quality}}       & \multicolumn{3}{c|}{\textit{Landmark Distance}} & \multicolumn{3}{c}{\textit{Temporal Coherence}} \\
\cline{2-6}
\cline{6-9}
\cline{10-12}
                        & FID $\downarrow$ &  L1 $\downarrow$ & SSIM $\uparrow$ & LPIPS $\downarrow$ & PSNR $\uparrow$ & Face $\downarrow$     & Body $\downarrow$    & Hand $\downarrow$    & FID-VID $\downarrow$       & FVD $\downarrow$ & L$_{wp}$ $\downarrow$       \\
\hline
DreamPose  & 66.77    &  12.67E-5  & 0.67 & 0.32 & 16.19 & 3.11 & 20.52 & 18.54 & 85.74 & 1083.82 & 4.56E-3 \\
MagicAnimate & 64.48    &  6.30E-5  & 0.76 & 0.21 & 20.09 & 3.85 & 20.49 & 9.94 & 28.03 & 600.65 & 3.82E-3\\
DisCo   &  67.37   &  6.72E-5  & 0.67 & 0.30 & 19.39 & 1.55 & 17.88 & 11.84 & 33.54 & 554.30 & 5.67E-3\\
AnimateAnyone  &  29.66   &   4.30E-5  & 0.80 & 0.16 & 21.76 & 1.83 & 14.47 & 8.06 & 21.59 & 416.90 & 3.65E-3 \\
Ours & \textbf{18.87}   &  \textbf{2.36E-5}  & \textbf{0.88} & \textbf{0.12} & \textbf{26.14} & \textbf{0.44} & \textbf{8.94} & \textbf{3.44} & \textbf{4.42} & \textbf{145.40} & \textbf{3.19E-3}\\
\hline
Make-Your-Anchor$^*$  & 25.68 &  6.64E-5 & 0.58 & 0.35 & 18.46  &  0.63 & 8.11 & 3.43 & 17.40 & 473.97 & 1.69E-3\\
Ours$^*$ &  \textbf{16.18} &  \textbf{2.61E-5} & \textbf{0.84} & \textbf{0.15} & \textbf{24.99} &   \textbf{0.52} & \textbf{7.12} & \textbf{3.36} & \textbf{4.87} & \textbf{161.24} & \textbf{1.14E-3}\\
\bottomrule
\end{tabular}
}
}
\label{tab:cmp_sota_metrics}
\end{table*}

\paragraph{Datasets.}
For our experiments, we collect suitable videos with centralized 2D speaking avatars from various sources, constituting a dataset of 11 subjects. All videos are segmented into clips of $\sim$10 seconds each.
Approximately 28 hours of footage featuring 5 identities, sourced from 
\cite{ahuja2020style} are used for pre-training. We randomly preserve 30 clips (300s) from each subject for evaluation. 
A generalized model is first trained on these 5 identities. 

To assess our method's capability for few-shot adaptation to unseen subjects, we further obtain 6 additional individuals for few-shot evaluations.
Two videos are sourced from ~\cite{ahuja2020style}, and four are self-recorded.
Each has a duration of 3-5 minutes. We split these videos into train and test sets, reserving 5 clips from each identity for testing while using the rest for training.

\paragraph{Implementation Details.}
We crop the avatar-centralized region from each clip according to the torso landmarks detected by DWPose~\cite{yang2023effective}. 
All video clips are pre-processed at a frame rate of 25 FPS with $512 \times 512$ resolution. 
The network of the Denosing Branch and Reference Branch is initialized from the UNet of Stable Diffusion~\cite{blattmann2023stable}.
The Motion Encoder consists of four convolution layers (kernel size: 4, stride: 2, using [16,32,96,256] channels) following SiLU activation. The spatial dimension of the extracted feature map is the same as the images’ appearance feature maps encoded from the VAE encoder. 
The Hand Feature Bank size $N_b$ is empirically set to 512.
During inference, we use a DDIM sampler for 50 denoising steps.

\paragraph{Comparison Methods.}
We compare our method with the most recent studies including \textbf{DreamPose}~\cite{karras2023dreampose}, \textbf{MagicAnimate}~\cite{xu2023magicanimate}, \textbf{DisCo}~\cite{wang2023disco}, \textbf{AnimateAnyone}~\cite{hu2023animate}, and \textbf{Make-Your-Anchor}~\cite{huang2024make}. 
We carefully conduct experiments using an open-source implementation of AnimateAnyone and the official implementations of other methods. 
The same training data used in our method is applied to finetune these methods, 
except for MagicAnimate and Make-Your-Anchor. 
For Make-Your-Anchor, we directly use its officially released checkpoint finetuned on the subject ``Seth'' from \cite{ahuja2020style}.

\begin{table*}[]
\centering
\caption{
\textbf{Ablations}. 
Lines in \textcolor{red}{red} color are evaluated on the test set of subject ``Oliver''~\cite{ahuja2020style} with various complex scenarios. 
Lines in \textcolor{blue}{blue} color are evaluated on the subject ``Prof''~\cite{ahuja2020style} test set, following finetuning with only limited data.
}
{\scriptsize
\resizebox{\linewidth}{!}{
\begin{tabular}{l|ccccc|ccc|ccc}
\toprule
\multirow{2}{*}{\begin{tabular}[c]{@{}l@{}}Experiment\\ (Oliver: \textcolor{red}{Red}, Prof: \textcolor{blue}{Blue})\end{tabular}}  
& \multicolumn{5}{c|}{\textit{Visual Quality}}       & \multicolumn{3}{c|}{\textit{Landmark Distance}} & \multicolumn{3}{c}{\textit{Temporal Coherence}} \\
\cline{2-6}
\cline{6-9}
\cline{10-12}
                        & FID $\downarrow$ &  L1 $\downarrow$ & SSIM $\uparrow$ & LPIPS $\downarrow$ & PSNR $\uparrow$ & Face $\downarrow$     & Body $\downarrow$    & Hand $\downarrow$    & FID-VID $\downarrow$       & FVD $\downarrow$ & L$_{wp}$ $\downarrow$      \\
\hline
\textcolor{red}{Ours w/o Ref Attn}&  53.45   & 4.45E-5 & 0.71 & 0.27 & 21.60 & 1.01 & 10.50 & 3.75 & 37.91 & 674.33 & 9.97E-4\\
\textcolor{red}{Ours w/o Textual Alignment} &  49.05   & 4.07E-5 & 0.70 & 0.23 & 22.33 & 0.64 & 7.29 & 2.90 & 13.51 & 402.18 & 8.99E-4\\
\textcolor{red}{Ours w/o Hand Attn} &  38.63  & 3.86E-5 & 0.72 & 0.21 & 22.20 & 0.59 & 8.01 & 4.67 & 10.55 & 317.78 & 9.38E-4\\
\textcolor{red}{Ours w/o 3D Guidance} &  48.28   &  4.14E-5 & 0.69 & 0.22 & 21.81 & 0.62 & 9.43 & 4.86 & 9.69 & 283.51 & 9.43E-4\\
\textcolor{red}{Ours} &  31.75  & 3.42E-5 & 0.75 & 0.18 & 23.18 & 0.51 & 6.37 & 2.21 & 6.47 & 169.37 & 7.67E-4\\
\hline
\textcolor{blue}{Finetune Data: 10s} &  17.09  &  3.04E-5 & 0.85 & 0.16 & 22.78 & 0.48 & 9.41 & 2.10 & 10.09 & 175.59 & 4.55E-3\\
\textcolor{blue}{Finetune Data: 30s} &  13.99  &  2.59E-5 & 0.86 & 0.14 & 23.90 & 0.44 & 8.48 & 1.72 & 7.68 & 144.48 & 4.36E-3\\
\textcolor{blue}{Finetune Data: 120s}& 13.32 &   2.42E-5 & 0.88 & 0.13 & 24.58 & 0.43 & 8.32 & 1.71 & 7.57 & 126.09 & 4.29E-3\\
\bottomrule
\end{tabular}
}
}
\label{tab:ablations}
\end{table*}
\begin{table}[]
\centering
\caption{\textbf{User Study}. Our user study follows the Mean Opinion Scores rating protocol. The rating score is from 1 to 5, higher is better.}
\resizebox{\linewidth}{!}{
\begin{tabular}{l|cccccc}
\toprule
Method $\rightarrow$ &
\begin{tabular}[c]{@{}c@{}}Dream\\ Pose\end{tabular} & 
\begin{tabular}[c]{@{}c@{}}Magic\\ Animate\end{tabular} & 
DisCo & 
\begin{tabular}[c]{@{}c@{}}Animate\\ Anyone\end{tabular} & 
\begin{tabular}[c]{@{}c@{}}Make-Your\\ -Anchor\end{tabular} & 
Ours \\
\hline
Movement Accuracy& 2.07 & 2.49 & 2.18 & 2.73 & 3.35 & 4.21 \\
Visual Quality& 1.53 & 2.16 & 1.89 & 2.31 & 2.88 & 4.10 \\
Temporal Coherence& 1.44 & 2.40 & 1.82 & 2.20 & 3.00 & 4.15 \\
\bottomrule
\end{tabular}
}
\label{tab:user_study}
\end{table}

\subsection{Comparison with State-of-the-Art Methods}

\paragraph{Quantitative Comparison.}
For quantitative evaluation, we employ comprehensive metrics from three perspectives: visual quality, landmark distance, and temporal coherence. 
To assess visual quality, we employ L1, FID~\cite{heusel2017gans}, SSIM~\cite{wang2004image}, LPIPS~\cite{zhang2018unreasonable}, and PSNR. In addition, we compute the landmarks on the face, body, and hands using OpenPose~\cite{cao2017realtime}, and report the absolute differences to indicate animation accuracy. For temporal coherence, we concatenate every consecutive 16 frames to form a sample, to report FID-VID~\cite{balaji2019conditional} and FVD~\cite{unterthiner2018towards} following~\cite{wang2023disco}. Moreover, we also report the flow warping error (L$_{wp}$) following~\cite{lai2018learning}.

The comparisons, evaluated on the 5 subjects used for pretraining, are presented in Table~\ref{tab:cmp_sota_metrics}.
Our method consistently achieves the best performance across all metrics. 
Note comparisons with Make-Your-Anchor are conducted using only the test set of subject ``Seth'' from~\cite{ahuja2020style}, as the authors provide results solely for this subject. 

\paragraph{Qualitative Comparison.}
Since our method pertains to video generation, subjective evaluation is essential for assessing its performance. We highly recommend readers watch our supplementary video for a comprehensive comparison with other methods.

We first present two samples including both self-driven and cross-driven results in Fig.~\ref{fig:cmp_sota}.
The two subjects are sourced from~\cite{ahuja2020no} and results are produced after few-show finetuning.
In the self-driven setting, driving signals are extracted from test videos of the same subject in the reference frame. For the cross-driven results, the generated video should preserve the identity and appearance features from the reference frame and precisely follow the body movements from the driving video. 
Notably, our method also supports high-quality facial expression transfer, a capability that other comparative methods lack.
From Fig.~\ref{fig:cmp_sota}, our approach showcases superior visual quality, particularly in areas such as hands and face. 
In addition, compared with the closest method \cite{huang2024make},
our approach surpasses the counterpart in terms of both visual quality and motion presentation. Note that the counterpart fails to accurately restore the suit color from the reference frame and introduces slight changes to the face shape.

\paragraph{User Study.}
We also conduct a user study for subjective visual evaluation. 25 participants are invited to rate each generated video from three perspectives.
1) Movement Accuracy: Does the generated video precisely follow the driving signal?
2) Visual Quality: How realistic is the generated video in terms of visual appearance?
3) Temporal Coherence: How consistent is the generated video in terms of temporal continuity?
For each option, a score is given from 1 to 5 (larger is better).
We prepare 10 sets of comparisons for all methods, resulting in a total of 250 questionnaires.
The averaged results are tabulated in Table~\ref{tab:user_study}. 
Our method obtains significantly higher scores in all three perspectives from users.

\begin{figure}[h]
\centering
\includegraphics[width=\linewidth]{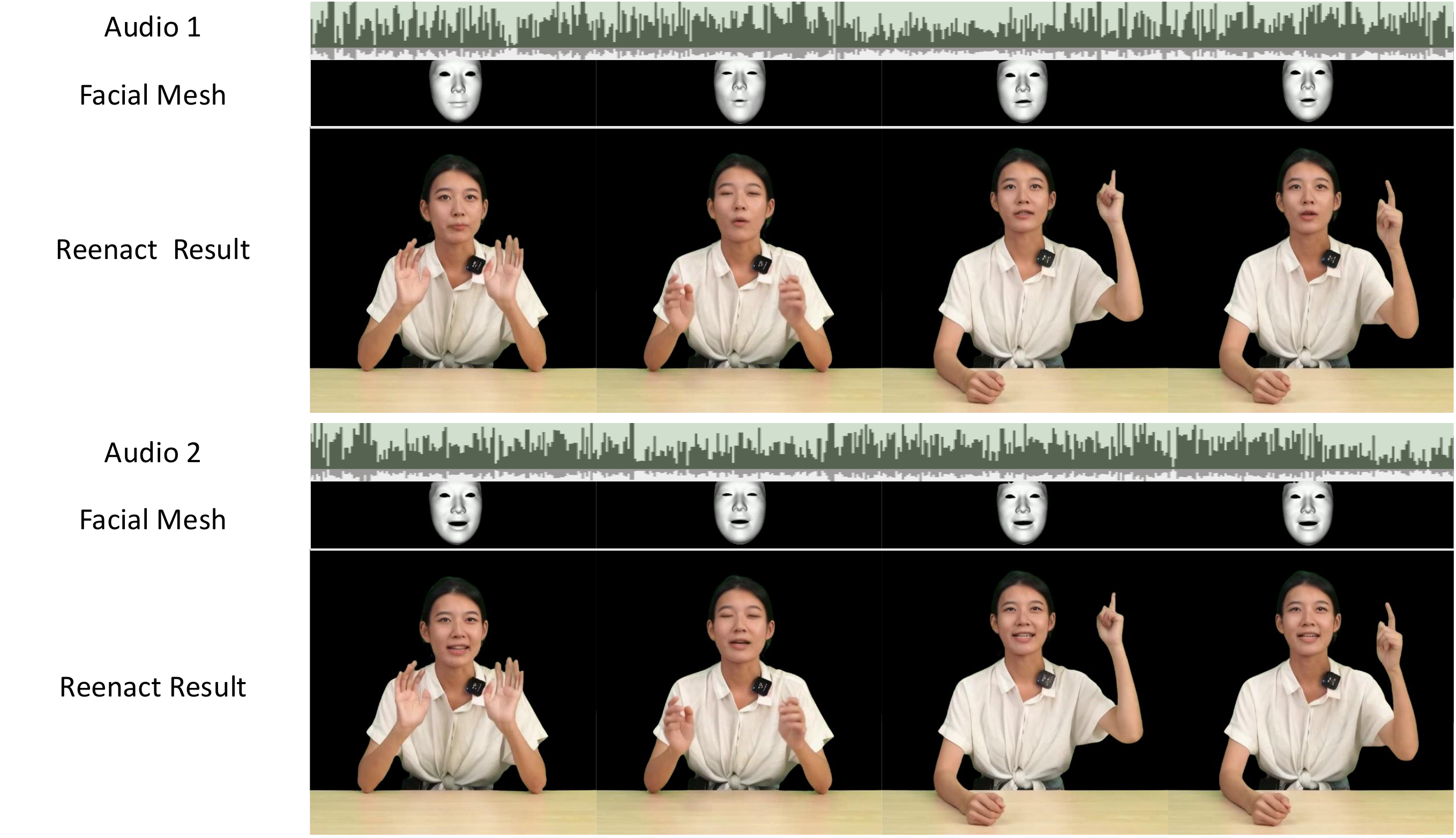}
\caption{
\textbf{Audio-Driven Facial Animation}. 
Facial expressions of reenacted results should match the facial meshes conditioned on the input audio.
}
\label{fig:wav2mesh}
\end{figure}

\paragraph{Combined with Audio-Driven Facial Animation.}
We further enhance the capabilities of the proposed framework by integrating an off-the-shelf audio-to-mesh method~\cite{fan2022faceformer} to directly enable audio-driven facial animation without any additional cost.
The results with limited data finetuning are shown in Fig.~\ref{fig:wav2mesh}. 
The facial meshes are generated conditioned on input audios. 
We show two cases with different audio inputs, where our method is able to generate accurate mouth shapes conditioned on given 3D meshes.

Additionally, we provide numerical comparison with popular works in lip-synchronization including Wav2Lip~\cite{prajwal2020lip}, ReTalking~\cite{cheng2022videoretalking}, SadTalker~\cite{zhang2023sadtalker}, and StyleSync~\cite{guan2023stylesync}. Results of SyncScore~\cite{chung2016out} are tabulated in Table~\ref{tab:cmp_sync}. 
Our method and ReTalking exhibit comparable performance, achieving SyncScores that are closest to the ground truth, providing superior realism in visualization.
\begin{table}[]
\centering
\caption{\textbf{Lip-Sync Comparisons}. We provide a numerical comparison of SyncScore~\cite{chung2016out} with popular works in lip-synchronization. $\Delta$SyncScore is the absolute difference between the ground truth (GT) and predictions.}
\resizebox{\linewidth}{!}{
\begin{tabular}{l|cccccc}
\toprule
Method $\rightarrow$ & Wav2Lip & SadTalker & ReTalking & StyleSync & Ours & GT   \\
\hline
SyncScore       & 8.21    & 2.11      & 6.50      & 7.45      & 6.08 & 6.31 \\
$\Delta$SyncScore & 1.9     & 4.2       & 0.19      & 1.14      & 0.23 & 0   \\
\bottomrule
\end{tabular}
}
\label{tab:cmp_sync}
\end{table}

\subsection{Ablation Studies}

\begin{figure}[h]
\centering
\includegraphics[width=\linewidth]{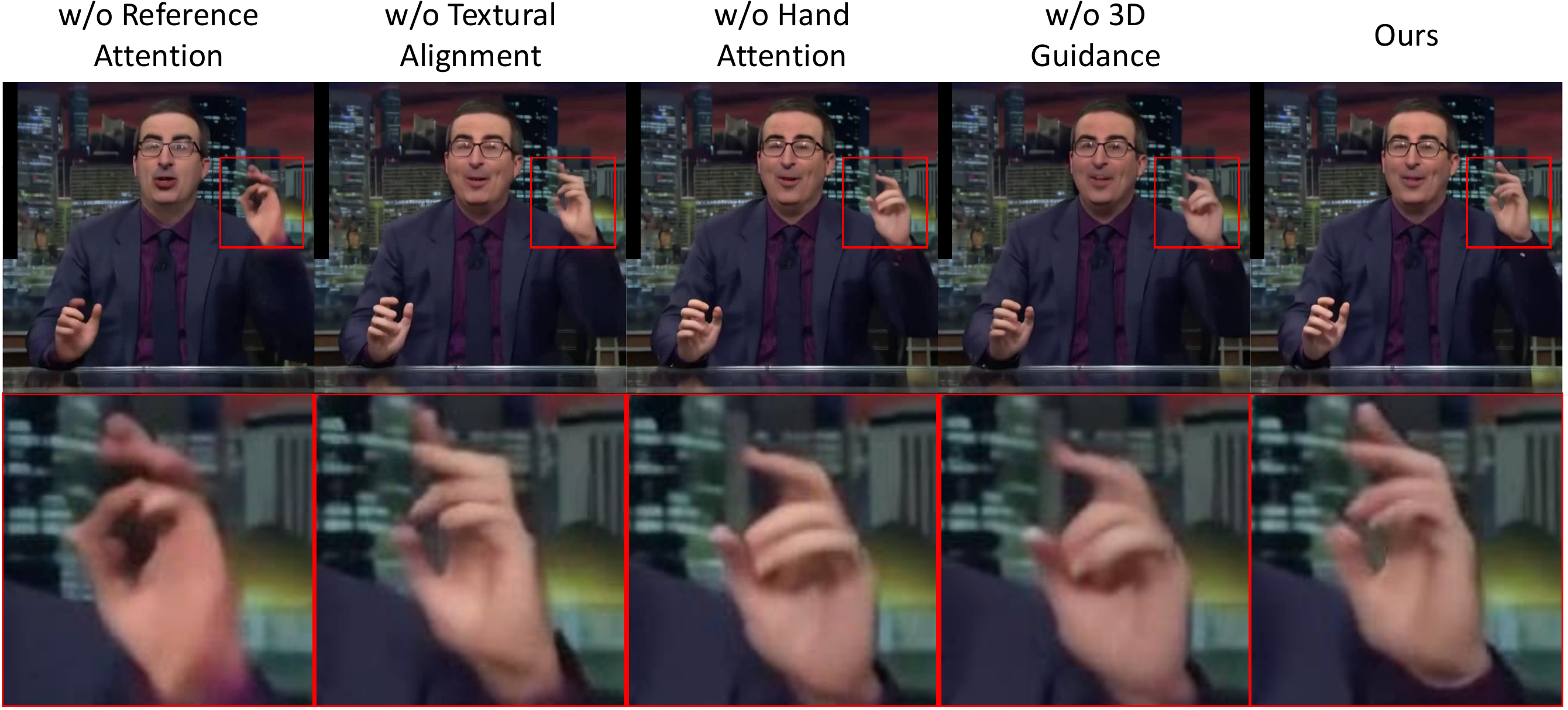}
\caption{
\textbf{Ablations}. 
Comparisons of several substitute designs.
The driven subject is from \copyright \emph{PATS}~\cite{ahuja2020style} (CC BY-NC 2.0)
}
\label{fig:ablations}
\end{figure}

\paragraph{Validation of Designs.}
To better demonstrate the effectiveness of our designs, we compare our method with several variants defined as follows: 
1) ``w/o Reference-Att'': We remove the reference attention layers from our Denoising Branch. 
2) ``w/o Textual Alignment'': We remove the proposed Motion-Enhanced Textural Alignment module.
3) ``w/o Hand Attn'': The hand attention layers are removed from the Denoising Branch.
4) ``w/o 3D Guidance'': We remove the overlay of the face meshes and hand meshes from our motion guidance, instead using face landmarks and hand outlines predicted by DWPose directly as driving signals.
Qualitative and quantitative results evaluated on the subject ``Oliver'' (from \cite{ahuja2020style}) are shown in Fig.~\ref{fig:ablations} and Table~\ref{tab:ablations} (red lines). 
Specifically, ``w/o Reference-Att'' leads to obvious artifacts across the whole frame, which also introduces noticeable temporal inconsistency even with finetuned temporal parameters. Indicated by the worst visual quality and temporal coherence in the table. 
``w/o Textual Alignment'' cannot stably restore the reference appearance when the body moves drastically. Blurred edges can be found in the results.
When removing the hand attention and the guidance of 3D meshes, detailed hand gestures cannot be expressed preciously, as indicated by the largest hand landmarks distance in the table.

\begin{figure}[h]
\centering
\includegraphics[width=\linewidth]{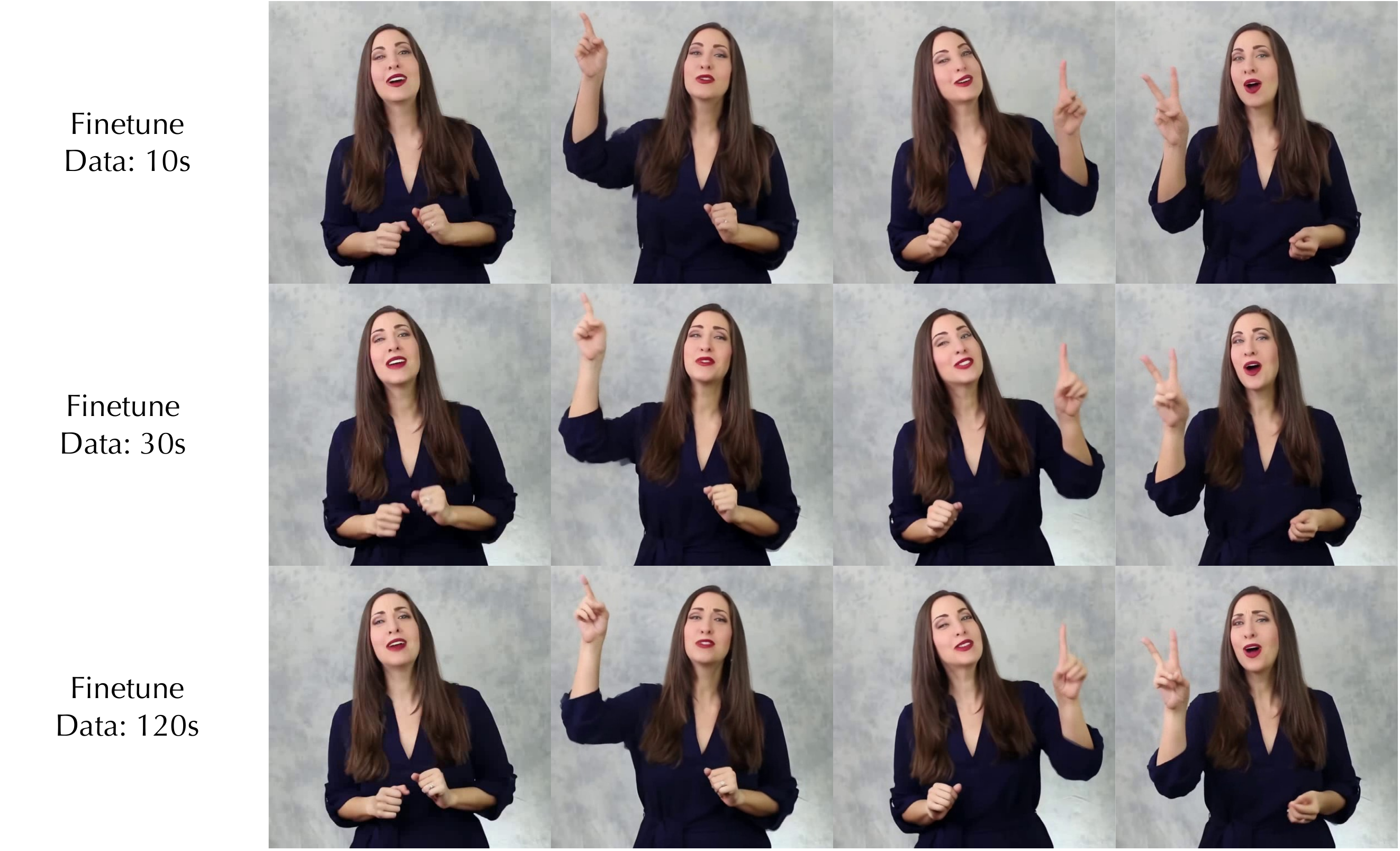}
\caption{
\textbf{Ablations}. 
Comparisons of different data durations for finetune.
The driven subject is from \copyright \emph{Vanessa Van Edwards, Science of People}.
}
\label{fig:finetune}
\end{figure}

\paragraph{Few-Shot Finetune.}
We further conduct experiments using different durations of finetune data. 
Both qualitative and quantitative results evaluated on the subject ``Prof'' are shown in Fig.~\ref{fig:finetune} and Table~\ref{tab:ablations} (blue lines). As shown in the figure, finetuning on 10s data can produce reasonably driven results, with only subtle artifacts around the body edges; there are hardly visual differences between results finetuned on 30s data and 120s data. Comparison results in Table~\ref{tab:ablations} also indicate that fine-tuning on 30s data already achieves commendable performance and providing more data only slightly improves the metric scores.

\section{Discussion and Conclusion}
\label{sec:conclusion}

\paragraph{Conclusion.}  
In this paper, we propose the \textbf{TALK-Act} framework which achieves high-fidelity 2D avatar reenactment with enhanced textural awareness with a diffusion model. We highlight several intriguing properties of our model: \textbf{1)} It synthesizes 2D speaking avatars with high fidelity and temporal consistency. The quality of our method clearly outperforms the previous state of the art. \textbf{2)} With an additional audio-to-mesh model, our method can produce lip-synced 2D avatars with desired gestures, which has rarely been achieved with a unified generative model. \textbf{3)} Our model creates high-fidelity results when tuned on subjects with only 30-second videos. Moreover, it still produces plausible results with 10 seconds of person-specific data.

\paragraph{Ethical Consideration.}
Our method demonstrates the capability to generate fabricated talks and speeches, which raises concerns about potential misuse. In light of this, we are committed to strictly controlling the distribution of our models and the generated content, limiting access exclusively to research purposes. 
On the other hand, significant progress has been made in detecting manipulations in facial regions, commonly referred to as deepfake detection~\cite{guan2022delving,Yao_2023_ICCV,oorloff2024avff,yan2024transcending}. However, detecting forgery in whole-body manipulations also requires urgent attention to prevent the malicious use of techniques similar to those proposed in this paper.

\begin{figure}[h]
\centering
\includegraphics[width=\linewidth]{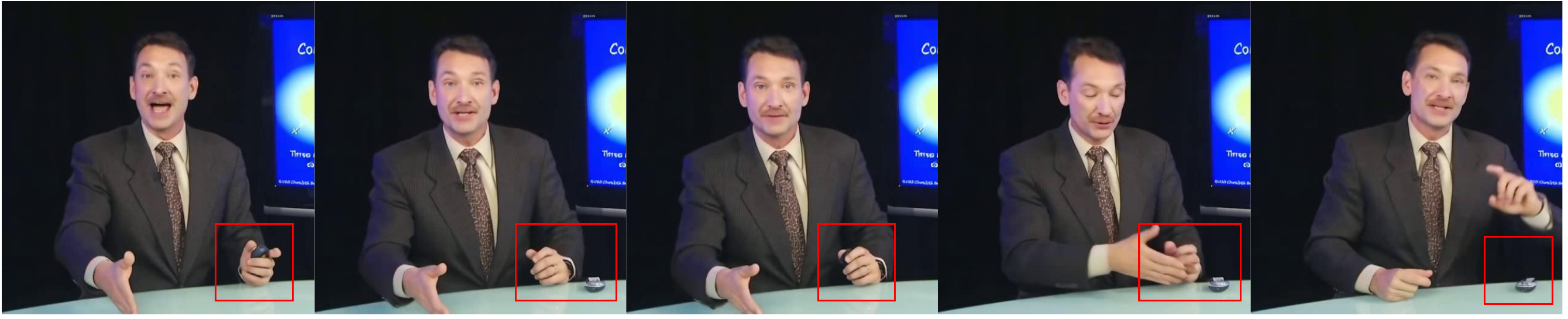}
\caption{
\textbf{Failures}. 
The driven avatar does not consistently perform interactions with the object in his hand.
The driven subject is from \copyright \emph{PATS}~\cite{ahuja2020style} (CC BY-NC 2.0)
}
\label{fig:failures}
\end{figure}

\paragraph{Limitation.} 
Unstable camera movements and background changes present challenges for our model. 
How to achieve 2D avatar reenactment under broader scenes is still worth exploring. 
A potential direction could involve disentangled modeling of body movements and background dynamics.
Meanwhile, our method is limited in its ability to consistently manage interactions between objects and the human body, as failure cases are shown in Fig.~\ref{fig:failures}.

\begin{acks}
This work is in part supported by National Natural Science Foundation of China with No. 62394322, Beijing Natural Science Foundation with No. L222024, as well as the Beijing National Research Center for Information Science and Technology (BNRist) key projects.
This study is also supported by the Ministry of Education, Singapore, under its MOE AcRF Tier 2 (MOET2EP20221- 0012), NTU NAP, and under the RIE2020 Industry Alignment Fund – Industry Collaboration Projects (IAF-ICP) Funding Initiative, as well as cash and in-kind contribution from the industry partner(s).
We would like to express our sincere gratitude to \textit{Charisma on Command Team} and \textit{Vanessa Van Edwards, Science of People Team} for granting us permission to use their video content in our research. Their contributions were invaluable in the development of this work.
\end{acks}

\bibliographystyle{ACM-Reference-Format}
\bibliography{main}

\clearpage

\end{document}